# FF-LINS: A Consistent Frame-to-Frame Solid-State-LiDAR-Inertial State Estimator

Hailiang Tang, Tisheng Zhang, Xiaoji Niu, Liqiang Wang, Linfu Wei, and Jingnan Liu

*Abstract*—Most of the existing LiDAR-inertial navigation systems are based on frame-to-map registrations, leading to inconsistency in state estimation. The newest solid-state LiDAR with a non-repetitive scanning pattern makes it possible to achieve a consistent LiDAR-inertial estimator by employing a frame-to-frame data association. In this letter, we propose a robust and consistent frame-to-frame LiDAR-inertial navigation system (FF-LINS) for solid-state LiDARs. With the INS-centric LiDAR frame processing, the keyframe point-cloud map is built using the accumulated point clouds to construct the frame-to-frame data association. The LiDAR frame-to-frame and the inertial measurement unit (IMU) preintegration measurements are tightly integrated using the factor graph optimization, with online calibration of the LiDAR-IMU extrinsic and time-delay parameters. The experiments on the public and private datasets demonstrate that the proposed FF-LINS achieves superior accuracy and robustness than the state-of-the-art systems. Besides, the LiDAR-IMU extrinsic and time-delay parameters are estimated effectively, and the online calibration notably improves the pose accuracy. The proposed FF-LINS and the employed datasets are open-sourced on GitHub (https://github.com/i2Nav-WHU/FF-LINS).

*Index Terms*—LiDAR-inertial navigation, state estimation, factor graph optimization, multi-sensor fusion navigation.

## I. INTRODUCTION

Light detection and ranging (LiDAR) navigation system has been widely used in navigation and mapping in this century. Conventionally, the iteration closest point (ICP)-based methods [1], [2] and the normal distributions transform (NDT)-based methods [3], [4] have been adopted for pose estimation, but they are mainly for dense point-cloud registration. The LiDAR sensors employed in autonomous vehicles and robots are commonly low-cost, and we can only obtain sparse point clouds from a LiDAR frame. Besides, initial pose estimation is also required to achieve successful iterations [5]. In addition, these methods are computationally intensive and may cost many computational resources. Due to these shortcomings, ICP-based and NDT-based methods are usually unsuitable for real-time navigation applications.

The real-time LiDAR odometry and mapping (LOAM) [6] is proposed without using the ICP or NDT. The edge and plane feature points are first extracted from a LiDAR frame by judging the smoothness of the local surface [6]. The LiDAR odometry is achieved by employing a frame-to-frame correspondence. The LiDAR frame is registered to the global feature map using the non-linear optimization method [7]. The feature-point detection and the frame-to-frame methods in LOAM [6] are mainly designed for the rotated 2-dimensional (2D) and 3-dimensional (3D) spinning LiDARs. LeGO-LOAM [8] further segments the ground plane points and adopts a two-step optimization, yielding higher computational performance.

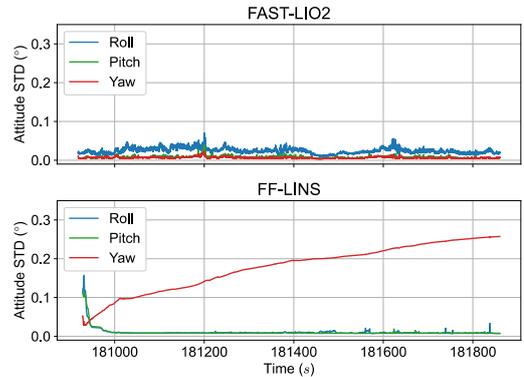

Fig. 1. Attitude standard deviation (STD) estimation comparison between FAST-LIO2 and FF-LINS on *Robot-campus* dataset. The yaw angle is unobservable for a LiDAR-inertial odometry, and thus the yaw STD will grow over time. FAST-LIO2 indicates inconsistent estimation, as the yaw STD does not grow. In contrast, FF-LINS exhibits great consistency in state estimation, because the yaw STD grows over time, while the roll and pitch STDs converge due to their observability.

The frame-to-map method may be meaningful in LiDAR mapping, as it can build a consistent map. However, without a prebuilt map, the LiDAR navigation system should be a dead-reckoning (DR) system, which may drift over time [9]. In contrast, the frame-to-map method constructs an absolute constraint between the current frame and the self-built map, resulting in inconsistency in pose estimation [10]. This problem may be more significant in multi-sensor fusion navigation with a tightly-coupled formulation, as it is impossible to incorporate absolute-positioning sensors, such as the global navigation satellite system (GNSS) [11].

The LiDAR point clouds are usually sampled at different times, which results in motion distortion. Hence, the micro-electro-mechanical system (MEMS) inertial measurement unit (IMU) can be employed to correct the distortion and construct a LiDAR-inertial navigation system. LOAM utilizes the

This research is partly funded by the National Key Research and Development Program of China (No. 2020YFB0505803) and the National Natural Science Foundation of China (No. 41974024). (*Corresponding authors: Tisheng Zhang; Xiaoji Niu.*)

Hailiang Tang, Liqiang Wang, and Linfu Wei are with the GNSS Research Center, Wuhan University, Wuhan 430079, China (e-mail: thl@whu.edu.cn; wlq@whu.edu.cn; weilf@whu.edu.cn).

Tisheng Zhang and Jingnan Liu are with the GNSS Research Center, Wuhan University, and the Hubei Luojia Laboratory, Wuhan 430079, China (e-mail: zts@whu.edu.cn; jnliu@whu.edu.cn).

Xiaoji Niu is with the GNSS Research Center, Wuhan University, Wuhan 430079, China, the Hubei Luojia Laboratory, Wuhan 430079, China, and the Artificial Intelligence Institute of Wuhan University, Wuhan 430079, China (email: xjniu@whu.edu.cn).

orientation and acceleration from an IMU to remove motion distortion, exhibiting improved accuracy [6]. In LIO-SAM, the IMU is applied in a LiDAR-inertial state estimator [12] within the framework of factor graph optimization (FGO) [13]. However, LIO-SAM [12] is a loosely-coupled system, as the LiDAR odometry is adopted in the state estimator rather than the raw LiDAR measurements. Besides, the LiDAR odometry in LIO-SAM is implemented by building a local point-cloud map, which is also inconsistent in state estimation. Hence, LIO-SAM has to employ a pose graph optimization [13] to fuse the LiDAR odometry and other absolute positioning sources, including the GNSS and the loop-closure constraint [12].

LINS [14] and FAST-LIO [15] are two similar tightly-coupled LiDAR-inertial odometry using the iterated extended Kalman filter (IEKF) [7]. The state estimation in [14], [15] is achieved by registering the extracted feature points in a LiDAR frame to the global feature-point map, which may result in inconsistency in the LiDAR-inertial state estimator. Specifically, the unobservable terms for a DR system, including the global yaw and the global position [10], can be wrongly observable by using the frame-to-map method, as depicted in Fig. 1. Moreover, the misalignment when registering the LiDAR frame to the global map may leading to inconsistent pose estimation relative to the IMU measurements. As a consequence, a wrong IMU biases estimation may occur and thus ruin the accuracy of the inertial navigation system (INS) [16]. In addition, the LiDAR-IMU extrinsic parameters cannot be effectively estimated with such a frame-to-map method. According to our experiments, the reason is that the system needs the extrinsic parameters to build the initial map; thus, the follow-up frame-to-map matching will prevent extrinsic parameters from being changed.

Recently, the newly solid-state LiDAR with a non-repetitive scanning pattern has been widely used for navigation and mapping [17]–[20]. LOAM-Livox employs the LOAM method for the solid-state LiDAR, Livox Mid-40, by adopting a new feature-extraction method [17]. For another solid-state LiDAR, Livox Horizon, with a different scanning pattern, LiLi-OM [18] proposes an applicable feature-extraction method. Besides, this method is integrated into a LiDAR-inertial scheme using a sliding-window optimization [18]. However, the frame-to-map method is employed in [18], and thus the inconsistent problem still exists. FAST-LIO2 [19] extends the work in FAST-LIO [15] by incorporating a direct registration method without feature extraction. Similarly, Faster-LIO [20] uses incremental voxels as the point-cloud spatial data structure rather than the incremental k-d tree in FAST-LIO2 [19]. Nevertheless, FAST-LIO2 and Faster-LIO adopt the same inconsistent state estimator.

As mentioned above, the LiDAR-inertial navigation system should be a DR system [9]. Hence, the LiDAR system should construct a relative constraint to achieve a consistent state estimation. LIPS designs a LiDAR-Inertial 3D Plane simultaneous-localization-and-mapping (SLAM) system with a robust relative plane anchor factor in graph-based optimization for indoor applications [21]. However, the planes should be segmented offline using the Point Cloud Library (PCL) [22], which cannot run in real-time. LIC-Fusion 2.0 [23] proposes a sliding-window plane-feature tracking method, which is then integrated into a multi-state constraint Kalman filter (MSCKF) [10]. As the relative constraints are constructed in LIPS and LIC-Fusion 2.0, the inconsistent problem in the state estimation should be solved. However, these methods are mainly designed for the 3D spinning LiDARs, and they are not applicable for solid-state LiDARs, such as Livox LiDARs with a non-repetitive scanning pattern. Besides, complex plane-extraction or plane-association algorithms should be employed to construct the relative constraints, and thus the computational complexity may significantly increase.

In this letter, we aim to construct a consistent solid-state-LiDAR-inertial navigation system (FF-LINS). We follow the INS-centric architecture in [11] to process the LiDAR data. A direct frame-to-frame data association algorithm is presented without explicitly extracting plane features. With the frame-to-frame association, a LiDAR frame-to-frame factor is proposed to construct a tightly-coupled LiDAR-inertial state estimator under the framework of FGO. The main contributions of our work are as follows:

● We propose a consistent solid-state-LiDAR-inertial state estimator that tightly integrates the LiDAR and IMU measurements within the FGO framework. The LiDAR-IMU extrinsic and time-delay parameters are all estimated and calibrated online to further improve the accuracy.

● A novel solid-state LiDAR frame-to-frame data association algorithm is presented. We build a direct keyframe point-cloud map with accumulated LiDAR frames with the prior INS poses. The data association is achieved by finding the nearest points in the keyframe point-cloud maps within the sliding window.

● A LiDAR frame-to-frame measurement model is proposed to achieve consistent state estimation in FGO. The LiDAR frame-to-frame measurement residuals, together with the Jacobians for the IMU poses and the LiDAR-IMU extrinsic parameters, are all analytically expressed.

● The proposed FF-LINS is comprehensively evaluated on both public and private datasets. The experiment results demonstrate that FF-LINS with the proposed consistent state estimator yields improved accuracy and robustness.

The remainder of this paper is organized as follows. We give an overview of the system pipeline in section II. The proposed frame-to-frame solid-state-LiDAR-inertial state estimator is presented in section III. The experiments and results are discussed in section IV for quantitative evaluation. Finally, we conclude the proposed FF-LINS.

## II. SYSTEM OVERVIEW

The system overview of the proposed FF-LINS is depicted in Fig. 2. The system pipeline is in an INS-centric architecture, and the proposed FGO is a sliding-window optimizer [11]. The INS is initialized firstly with zero position and zero yaw angle, while the roll and pitch angles are determined from the accelerometer measurements [16]. We can also obtain a rough gyroscope biases estimation [11] if zero-velocity conditions are detected. Once the system is initialized, the INS mechanization is conducted to provide prior poses for LiDAR frame processing. With the prior INS poses, the LiDAR frame is processed, and the LiDAR keyframe is selected. Then, we build the keyframe point-cloud map with accumulated LiDAR frames. Thus, the frame-to-frame data association can be conducted by finding the nearest points in all keyframe point-cloud maps within the sliding window. Finally, the LiDAR frame-to-frame factor can be constructed between the LiDAR keyframes. The LiDAR and IMU measurements are tightly coupled within the FGO framework to perform the maximum-a-posterior estimation [7].

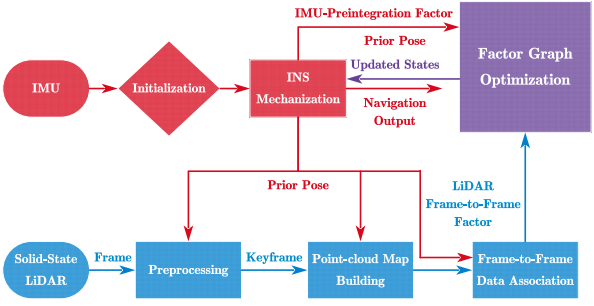

Fig. 2. System overview of the proposed FF-LINS.

## III. METHODOLOGY

The proposed frame-to-frame solid-state-LiDAR-inertial navigation system is presented in this section. We will first introduce the INS-centric LiDAR frame processing. Then, the direct frame-to-frame data association algorithm is proposed. Finally, we present the consistent state estimator with the analytical form of the frame-to-frame measurement residuals and Jacobians.

### A. INS-Centric LiDAR Frame Processing

We follow the INS-centric processing architecture [11] to process the LiDAR frame. The high-frequency INS poses will be employed to assist the LiDAR frame processing, including the motion-distortion compensation, the keyframe selection, and the keyframe point-cloud map building.

#### 1) INS Mechanization

Once the INS is initialized, the INS mechanization is conducted. We adopt a reduced INS kinematic model [24] in the proposed FF-LINS, and it can be written as

$$
\begin{aligned}
\dot{\boldsymbol{p}}_{\mathrm{wb}}^{\mathrm{w}} &= \boldsymbol{v}_{\mathrm{wb}}^{\mathrm{w}}, \\
\dot{\boldsymbol{v}}_{\mathrm{wb}}^{\mathrm{w}} &= \mathbf{R}_{\mathrm{b}}^{\mathrm{w}} \boldsymbol{f}^{\mathrm{b}} + \boldsymbol{g}^{\mathrm{w}}, \\
\dot{\mathbf{q}}_{\mathrm{b}}^{\mathrm{w}} &= \frac{1}{2} \mathbf{q}_{\mathrm{b}}^{\mathrm{w}} \otimes \begin{bmatrix} 0 \\ \boldsymbol{w}^{\mathrm{b}} \end{bmatrix},
\end{aligned} \tag{1}
$$

where $\boldsymbol{p}_{\mathrm{wb}}^{\mathrm{w}}$ and $\boldsymbol{v}_{\mathrm{wb}}^{\mathrm{w}}$ are the position and velocity of the IMU frame (b-frame) in the world frame (w-frame), respectively; the quaternion $\mathbf{q}_{\mathrm{b}}^{\mathrm{w}}$ and the rotation matrix $\mathbf{R}_{\mathrm{b}}^{\mathrm{w}}$ denote the rotation of the b-frame with respect to the w-frame; $\boldsymbol{g}^{\mathrm{w}}$ is the gravity vector in the w-frame; $\boldsymbol{w}^{\mathrm{b}}$ and $\boldsymbol{f}^{\mathrm{b}}$ are the compensated angular velocity and acceleration from the IMU, respectively; $\otimes$ denotes the quaternion product. The IMU frame is defined as the IMU body frame, *i.e.* the front-right-down frame. The w-frame is defined at the initial point with zero position and zero yaw angle, while the roll and pitch angle are gravity-aligned [16]. The INS mechanization can be formulated by adopting the kinematic model in (1) to obtain high-frequency INS poses.

#### 2) LiDAR Frame Preprocessing

A keyframe-based LiDAR frame processing is employed in the proposed FF-LINS. When a new LiDAR frame is valid, we preprocess the LiDAR frame with the prior INS pose. Specifically, the interpolated INS poses are adopted to retrieve undistorted point clouds [25]. The direct point-cloud processing method has been proven to be more robust than the feature-based methods while achieving the same accuracy [19]. Hence, the direct-based method is employed without explicitly extracting plane features. The undistorted point clouds of a LiDAR frame are directly downsampled using a voxel grid filter [22], and the leaf size is set to 0.5 m [19].

In the proposed INS-centric architecture, the state-estimation update will only be conducted when a new LiDAR keyframe is selected, and the INS can output continuous poses during the period [11], as depicted in Fig. 2. In other words, only the LiDAR keyframe will be adopted to perform state estimation. The proposed INS-centric processing can significantly save computational costs and thus improve real-time performance without decreasing accuracy.

To fully use the short-time accuracy of the INS, the LiDAR keyframe should be selected within a short interval. Besides, we should also consider LiDAR's motions to build up valid frame-to-frame LiDAR data association. If the translation or the rotation change exceeds thresholds [12], *e.g.* 0.4 m and 10 °, a new LiDAR keyframe will be selected. Here, the translation and the rotation are derived from the prior INS poses. If the motion of the LiDAR is small for a long interval, *e.g.* 0.5 s, we will also pick up a keyframe. When a new LiDAR keyframe is selected, the frame-to-frame data association can be conducted to perform the consistent state estimation. Nevertheless, the LiDAR non-keyframes will be reserved to build the point-cloud map corresponding to the new LiDAR keyframe.

#### 3) Point-Cloud Map Building

The point clouds of a single solid-state-LiDAR are sparse [17], which is inconducive for the frame-to-frame data association. Relatively dense point clouds can be obtained by accumulating several LiDAR frames due to the non-repetitive scanning pattern of the solid-state LiDAR [25]. Hence, it is convenient to construct the frame-to-frame data association with such dense point clouds. As we can obtain high-accuracy poses from the INS in a short time, the LiDAR frames can be further accumulated with the prior INS poses. Specifically, all LiDAR frames since the previous keyframe, including the non-keyframes and the new keyframe, will be employed together to build the point-cloud map corresponding to the new keyframe. As depicted in Fig. 3, the point clouds in LiDAR non-keyframes will be projected to the corresponding time of the LiDAR keyframe with the prior poses from the INS. Finally, we obtain the LiDAR keyframe point-cloud map $\mathbf{M}$, which will be adopted for the frame-to-frame data association. The keyframe point-cloud map is also downsampled using the voxel grid filter in section III.A.2.

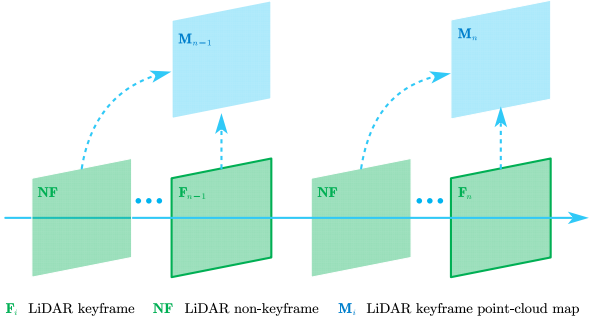

F<sub> </sub> LiDAR keyframe    NF LiDAR non-keyframe    $\mathbf{M}_i$ LiDAR keyframe point-cloud map

Fig. 3. An illustration of the LiDAR keyframe and keyframe point-cloud map.

## B. Frame-to-Frame Data Association

The frame-to-frame data association can be carried on with the keyframe point-cloud map in section III.A.3. As mentioned above, only the LiDAR keyframe will be adopted for state estimation. Specifically, only the point clouds in the LiDAR keyframe will be employed to construct the frame-to-frame data association. In other words, the keyframe point-cloud map $\mathbf{M}$ covers much more fields of view than the LiDAR keyframe point cloud $\mathbf{F}$. That is the reason that we can build up valid frame-to-frame data associations.

An illustration of the frame-to-frame data association is depicted in Fig. 4. Considering that we have $n+1$ LiDAR keyframes in the sliding window, we will associate the latest keyframe point cloud $\mathbf{F}_n$ with the keyframe point-cloud maps $\mathbf{M}_i, i \in [0, n-1]$. For a point $\boldsymbol{p}^{r_i}$ in $\mathbf{F}_n$, where $\mathrm{r}$ denotes the LiDAR frame (r-frame), it can be projected to the keyframe point-cloud maps with the prior LiDAR pose $\left\{ \boldsymbol{p}^{w}_{wr_n}, \mathbf{q}^{w}_{r_n} \right\}$ from the INS and the estimated LiDAR pose $\left\{ \boldsymbol{p}^{w}_{wr_i}, \mathbf{q}^{w}_{r_i} \right\}$. As shown in Fig. 4, the projection of the point $\boldsymbol{p}^{r_i}$ in $\mathbf{M}_i$ can be written as

$$\boldsymbol{p}^{r_i} = \left( \mathbf{R}^{w}_{r_i} \right)^{T} \left( \mathbf{R}^{w}_{r_n} \boldsymbol{p}^{r_n} + \boldsymbol{p}^{w}_{wr_n} - \boldsymbol{p}^{w}_{wr_i} \right). \tag{2}$$

In the proposed FF-LINS, the frame-to-frame association is equal to the direct plane-point registration [19], and we treat all point clouds as plane-point candidates. With the projected point $\boldsymbol{p}^{r_i}$ (the red points in Fig. 4), we find its five nearest points $\boldsymbol{p}_e, e \in [1,5]$ (the green points in Fig. 4) in the keyframe point-cloud map $\mathbf{M}_i$. An overdetermined linear equation [25] can be constructed using the five nearest points to solve the following plane equation as

$$\boldsymbol{n}^{T} \boldsymbol{p} + d = 0, \tag{3}$$

where $\boldsymbol{p}$ is a point on the plane; $\boldsymbol{n}$ is the normalized normal vector of the plane; $d$ is a distance that satisfies the equation (3). The fitted local plane is checked by calculating the point-to-plane distance as

$$\mathrm{dis}_{\boldsymbol{p}_e} = \left| \boldsymbol{n}^{T} \boldsymbol{p}_e + d \right|, e \in [1,5]. \tag{4}$$

If $\mathrm{dis}_{\boldsymbol{p}_e} < 0.1m$ for all the five points [25], the fitted plane will be used for the following processing. Otherwise, the frame-to-frame association for the point $\boldsymbol{p}^{r_i}$ in $\mathbf{M}_i$ is failed. A similar method in [6], [19] is used to check the $\mathrm{dis}_{\boldsymbol{p}^{r_i}} = \left| \boldsymbol{n}^{T} \boldsymbol{p}^{r_i} + d \right|$ to validate the frame-to-frame association.

Finally, we obtain the frame-to-frame associations between the latest LiDAR keyframe and other keyframes in the sliding window. The fitted plane parameters $\{ \boldsymbol{n}, d \}$ for each frame-to-frame association will be employed to construct a LiDAR frame-to-frame measurement in FGO. Hence, we can build relative measurements between the latest LiDAR keyframe and other keyframes to achieve a consistent state estimation.

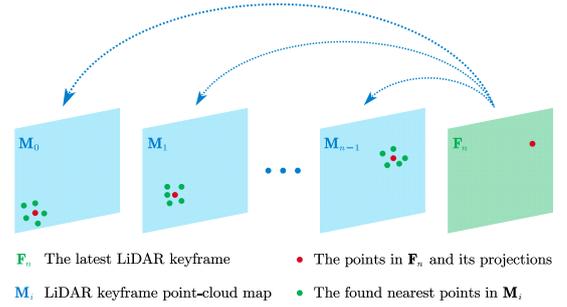

F<sub>n</sub> The latest LiDAR keyframe    • The points in $\mathbf{F}_n$ and its projections

$\mathbf{M}_i$ LiDAR keyframe point-cloud map    • The found nearest points in $\mathbf{M}_i$

Fig. 4. An illustration of the frame-to-frame data association.

## C. Factor Graph Optimization

The INS information is fully utilized in the INS-centric LiDAR frame processing, and thus we obtain the undistorted LiDAR keyframe and the keyframe point-cloud map. The frame-to-frame data association is achieved by constructing plane measurements between the latest LiDAR keyframe and the keyframe point-cloud maps. Hence, the consistent LiDAR-inertial state estimator is achieved by tightly integrating the LiDAR frame-to-frame and IMU preintegration measurements within the FGO framework.

### 1) Formulation

The proposed consistent state estimator is a sliding-window optimizer. The state vector $\boldsymbol{X}$ in FF-LINS can be defined as follows

$$\begin{aligned}
\boldsymbol{X} &= \left[ \boldsymbol{x}_0, \boldsymbol{x}_1, ..., \boldsymbol{x}_n, \boldsymbol{x}^{b}_r, t_d \right], \\
\boldsymbol{x}_k &= \left[ \boldsymbol{p}^{w}_{wb_k}, \mathbf{q}^{w}_{b_k}, \boldsymbol{v}^{w}_{wb_k}, \boldsymbol{b}_{g_k}, \boldsymbol{b}_{a_k} \right], k \in [0, n], \\
\boldsymbol{x}^{b}_r &= \left[ \boldsymbol{p}^{b}_{br}, \mathbf{q}^{b}_r \right],
\end{aligned} \tag{5}$$

where $\boldsymbol{x}_k$ is the IMU state at each time node, including the position, the attitude quaternion, and the velocity in the w-frame, and the gyroscope biases $\boldsymbol{b}_g$ and the accelerometer biases $\boldsymbol{b}_a$; $n$ is the size of the sliding window, $i.e.$ the number of the IMU preintegration factors; $\boldsymbol{x}^{b}_r$ is the LiDAR-IMU extrinsic parameters; $t_d$ denotes the time delay between the LiDAR and the IMU data.

The state estimation is conducted by solving the following non-linear least squares problems of the form

$$\min_{\boldsymbol{X}} \left\{ \begin{array}{c} \sum_{h \in [0, m]} \left\| \mathbf{r}_R \left( \tilde{\boldsymbol{z}}^{R}_h, \boldsymbol{X} \right) \right\|^2_{\boldsymbol{\Sigma}^{R}_h} + \\ \sum_{k \in [1, n]} \left\| \mathbf{r}_{Pre} \left( \tilde{\boldsymbol{z}}^{Pre}_{k-1,k}, \boldsymbol{X} \right) \right\|^2_{\boldsymbol{\Sigma}^{Pre}_{k-1,k}} + \left\| \mathbf{r}_p - \mathbf{H}_p \boldsymbol{X} \right\|^2 \end{array} \right\}, \tag{6}$$

where $\boldsymbol{r}_{R}$ are the residuals for the LiDAR frame-to-frame measurements, which construct relative pose constraints between the latest LiDAR keyframe and other keyframes in the sliding window; $m$ denotes the total number of the LiDAR measurements; $\boldsymbol{r}_{Pre}$ are the residuals for the IMU preintegration measurements [24]; $\{\boldsymbol{r}_{p}, \mathbf{H}_{p}\}$ denote the prior information from the marginalization [7]. We adopt the Levenberg-Marquardt algorithm in Ceres solver [26] to solve the non-linear least squares problem in (6).

### 2) Frame-to-Frame Measurement Residuals

The LiDAR frame-to-frame measurement residual is equal to the point-to-plane distance [18]. Nevertheless, the proposed frame-to-frame measurement model is consistent, while the frame-to-map measurement model in the existing works is inconsistent, such as [12], [14], [18]–[20]. The LiDAR-inertial extrinsic parameters $\{\boldsymbol{p}_{bl}^{b}, \mathbf{q}_{l}^{b}\}$ and the time delay $t_{d}$ are all incorporated into the frame-to-frame measurement model for online estimation and calibration. For convenience, the time delay $t_{d}$ will be omitted in the following parts, and we can refer to [23] for further details.

Suppose the raw point $\boldsymbol{p}^{\tau_{n,h}}$ in $\mathbf{F}_{n}$ is associated in the keyframe point-cloud map $\mathbf{M}_{i}$. Then, we have the associated plane parameters $\{\tilde{\boldsymbol{n}}_{h}, \tilde{d}_{h}\}$. The residual of the LiDAR frame-to-frame measurement $h$ is the function of the IMU poses $\{\boldsymbol{p}_{wb_{n}}^{w}, \mathbf{q}_{b_{n}}^{w}\}$ and $\{\boldsymbol{p}_{wb_{i}}^{w}, \mathbf{q}_{b_{i}}^{w}\}$, and the LiDAR-IMU extrinsic parameters $\{\boldsymbol{p}_{bl}^{b}, \mathbf{q}_{l}^{b}\}$. Consequently, the LiDAR frame-to-frame measurement residual can be written as

$$
\begin{aligned}
\boldsymbol{r}_{R}\left(\tilde{\boldsymbol{z}}_{h}^{R}, \boldsymbol{X}\right) &= \left(\tilde{\boldsymbol{n}}_{h}\right)^{T} \boldsymbol{p}^{\tau_{i,h}} + \tilde{d}_{h}, \\
\boldsymbol{p}^{\tau_{i,h}} &= \left(\mathbf{R}_{l}^{b}\right)^{T}\left(\boldsymbol{p}^{b_{i,h}} - \boldsymbol{p}_{bl}^{b}\right), \\
\boldsymbol{p}^{b_{i,h}} &= \left(\mathbf{R}_{b_{i}}^{w}\right)^{T}\left(\boldsymbol{p}^{w_{h}} - \boldsymbol{p}_{wb_{i}}^{w}\right), \\
\boldsymbol{p}^{w_{h}} &= \mathbf{R}_{b_{n}}^{w} \boldsymbol{p}^{b_{n,h}} + \boldsymbol{p}_{wb_{n}}^{w}, \\
\boldsymbol{p}^{b_{n,h}} &= \mathbf{R}_{l}^{b} \boldsymbol{p}^{\tau_{n,h}} + \boldsymbol{p}_{bl}^{b},
\end{aligned} \tag{7}
$$

where $\boldsymbol{p}^{\tau_{i,h}}$ is the projection of the raw point $\boldsymbol{p}^{\tau_{n,h}}$ in the r-frame of the keyframe $i$; $\boldsymbol{p}^{w_{h}}$ denotes the projection in the w-frame; $\boldsymbol{p}^{b_{i,h}}$ and $\boldsymbol{p}^{b_{n,h}}$ denote the projections in the b-frame corresponding to the LiDAR keyframe $i$ and $n$. The LiDAR frame-to-frame measurement model in (7) is similar to the visual reprojection model in visual navigation [11], and they are all relative-pose measurement models. In other words, the proposed LiDAR frame-to-frame measurement model is consistent in terms of state estimation.

As the direct method without explicitly extracting the plane feature is adopted, the covariance $\boldsymbol{\Sigma}_{h}^{R}$ may be difficult to be determined. Thanks to the frame-to-frame association method in FF-LINS, the covariance $\boldsymbol{\Sigma}_{h}^{R}$ can be obtained offline by error statistics. Specifically, we can first generate the keyframe point clouds $\mathbf{F}$ and keyframe point-cloud maps $\mathbf{M}$ with FF-LINS. Then, the ground-truth poses can be employed to build the frame-to-frame association presented in section III.B. The frame-to-frame measurement errors can also be calculated using the ground-truth poses. Finally, we can analyze the distributions of all the frame-to-frame measurement errors to obtain the covariance. According to our experiments, the covariance can be set as $\boldsymbol{\Sigma}_{h}^{R} = \sigma^{2} \mathbf{I}$, where $\sigma$ is about 0.1 m.

### 3) Jacobians of the Frame-to-Frame Measurement Residual

Using the error-perturbation method in [24], we can obtain the analytical Jacobians of $\boldsymbol{r}_{R}$ in (7) w.r.t the pose errors $\{\delta \boldsymbol{p}_{wb_{n}}^{w}, \delta \boldsymbol{\phi}_{wb_{n}}^{w}\}$ and $\{\delta \boldsymbol{p}_{wb_{i}}^{w}, \delta \boldsymbol{\phi}_{wb_{i}}^{w}\}$, and the LiDAR-IMU extrinsic errors $\{\delta \boldsymbol{p}_{bl}^{b}, \delta \boldsymbol{\phi}_{l}^{b}\}$. Here, $\boldsymbol{\phi}$ denotes the rotation vector of a quaternion $\mathbf{q}$, and $\delta \boldsymbol{\phi}$ represents the attitude errors [24]. Specifically, the Jacobians w.r.t the pose errors $\{\delta \boldsymbol{p}_{wb_{n}}^{w}, \delta \boldsymbol{\phi}_{wb_{n}}^{w}\}$ can be formulated as

$$
\left\{
\begin{aligned}
\frac{\partial \boldsymbol{r}_{R}}{\partial \delta \boldsymbol{p}_{wb_{n}}^{w}} &= \left(\tilde{\boldsymbol{n}}_{h}\right)^{T}\left(\mathbf{R}_{l}^{b}\right)^{T}\left(\mathbf{R}_{b_{i}}^{w}\right)^{T} \\
\frac{\partial \boldsymbol{r}_{R}}{\partial \delta \boldsymbol{\phi}_{wb_{n}}^{w}} &= -\left(\tilde{\boldsymbol{n}}_{h}\right)^{T}\left(\mathbf{R}_{l}^{b}\right)^{T}\left(\mathbf{R}_{b_{i}}^{w}\right)^{T} \mathbf{R}_{b_{n}}^{w}\left[\boldsymbol{p}^{b_{n,h}}\right]_{\times}
\end{aligned}
\right. , \tag{8}
$$

where $[\bullet]_{\times}$ denotes the skew-symmetric matrix of a vector [16]; $\boldsymbol{p}^{b_{n,h}}$ is the point projection in (7). Similarly, the Jacobians w.r.t the pose errors $\{\delta \boldsymbol{p}_{wb_{i}}^{w}, \delta \boldsymbol{\phi}_{wb_{i}}^{w}\}$ can be written as

$$
\left\{
\begin{aligned}
\frac{\partial \boldsymbol{r}_{R}}{\partial \delta \boldsymbol{p}_{wb_{i}}^{w}} &= -\left(\tilde{\boldsymbol{n}}_{h}\right)^{T}\left(\mathbf{R}_{l}^{b}\right)^{T}\left(\mathbf{R}_{b_{i}}^{w}\right)^{T} \\
\frac{\partial \boldsymbol{r}_{R}}{\partial \delta \boldsymbol{\phi}_{wb_{i}}^{w}} &= \left(\tilde{\boldsymbol{n}}_{h}\right)^{T}\left(\mathbf{R}_{l}^{b}\right)^{T}\left[\left(\mathbf{R}_{b_{i}}^{w}\right)^{T}\left(\boldsymbol{p}^{w_{h}} - \boldsymbol{p}_{wb_{i}}^{w}\right)\right]_{\times}
\end{aligned}
\right. , \tag{9}
$$

where $\boldsymbol{p}^{w_{h}}$ is the point projection in (7). We can also obtain the Jacobians w.r.t the LiDAR-IMU extrinsic errors $\{\delta \boldsymbol{p}_{bl}^{b}, \delta \boldsymbol{\phi}_{l}^{b}\}$ as

$$
\left\{
\begin{aligned}
\frac{\partial \boldsymbol{r}_{R}}{\partial \delta \boldsymbol{p}_{bl}^{b}} &= \left(\tilde{\boldsymbol{n}}_{h}\right)^{T}\left(\mathbf{R}_{b_{n,i}}^{r} - \left(\mathbf{R}_{l}^{b}\right)^{T}\right) \\
\frac{\partial \boldsymbol{r}_{R}}{\partial \delta \boldsymbol{\phi}_{l}^{b}} &= \left(\tilde{\boldsymbol{n}}_{h}\right)^{T}\left(\left[\left(\mathbf{R}_{l}^{b}\right)^{T}\left(\boldsymbol{p}^{b_{i,h}} - \boldsymbol{p}_{bl}^{b}\right)\right]_{\times} - \mathbf{R}_{b_{n,i}}^{r} \mathbf{R}_{l}^{b}\left[\boldsymbol{p}^{\tau_{n,h}}\right]_{\times}\right)
\end{aligned}
\right. , \tag{10}
$$

where $\boldsymbol{p}^{b_{i,h}}$ is the point projection in (7), and $\boldsymbol{p}^{\tau_{n,h}}$ is the raw point in the keyframe $\mathbf{F}_{n}$. The rotation matrix $\mathbf{R}_{b_{n,i}}^{r}$ can be written as

$$
\mathbf{R}_{b_{n,i}}^{r} = \left(\mathbf{R}_{l}^{b}\right)^{T}\left(\mathbf{R}_{b_{i}}^{w}\right)^{T} \mathbf{R}_{b_{n}}^{w}. \tag{11}
$$

Finally, we obtain the Jacobians of $\boldsymbol{r}_{R}$ w.r.t pose errors and the LiDAR-IMU extrinsic errors in (8), (9), and (10), which are all analytically expressed.

### 4) Outlier Culling

We adopt a two-step optimization when solving the non-linear least squares problems in (6). As wrong frame-to-frame associations may occur, especially in complex environments, we employ the Huber robust cost function to reduce the impacts of the outliers. After the first optimization, a chi-square test is employed to determine and remove the LiDAR frame-to-frame factor outliers from the optimizer. The estimated states will be further optimized in the second optimization. We do not need to remove outliers after the two-step optimization because new frame-to-frame associations will always be constructed when a new LiDAR keyframe is selected. All in all, the outlier-culling methods in FF-LINS can effectively improve the robustness in complex environments.

## IV. EXPERIMENTS AND RESULTS

In this section, we conduct exhaustive experiments to evaluate the proposed FF-LINS. The public and private datasets are all employed to examine the accuracy and robustness of FF-LINS. The running-time analysis is also conducted to evaluate to real-time performance of FF-LINS.

### A. Datasets and Implementation

The employed public datasets are the *Lili-OM* [18] and *R3LIVE* datasets [27]. The *LiLi-OM* dataset includes the Livox Horizon and the built-in IMU, and the three longest sequences are adopted, including the sequences *Schloss-1, Schloss-2*, and *East*. The *R3LIVE* dataset includes the Livox AVIA and the built-in IMU, and the three longest sequences with end-to-end trajectories are adopted, including the sequences *hku_main_building*, *hkust_campus_00*, and *hkust_campus_01*.

The private datasets are collected with a low-speed wheeled robot with an average speed of around 1.5 m/s. The sensors include a solid-state LiDAR (Livox Mid-70 with a frame rate of 10 Hz) and an industrial-grade MEMS IMU (ADI ADIS16465 with a gyroscope bias instability of 2 °/hr and a frame rate of 200 Hz), as depicted in Fig. 5. The solid-state LiDAR and the IMU are well-synchronized through hardware triggers. The ground-truth system is a high-accuracy GNSS/INS integrated navigation system using the GNSS-RTK and a navigation-grade IMU [16]. Besides, the ground truth (0.02 m for position and 0.01 deg for attitude) is generated by post-processing software. As depicted in Fig. 6, there are four sequences in the *Robot* dataset, including *campus* (1.33 km and 934 s), *building* (2.56 km and 1825 s), *playground* (1.33 km and 969 s), and *park* (1.46 km and 1326 s). The testing scenes contain various structured and unstructured environments, as shown in Fig. 6. Besides, many moving objects, such as pedestrians, bicycles, and vehicles, make it great challenging to achieve robust navigation.

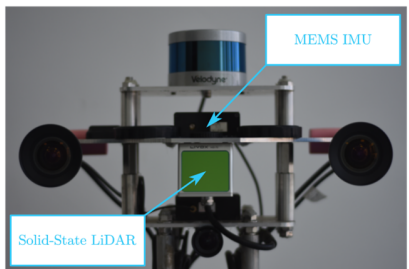

Fig. 5. Equipment setup in the *Robot* dataset.

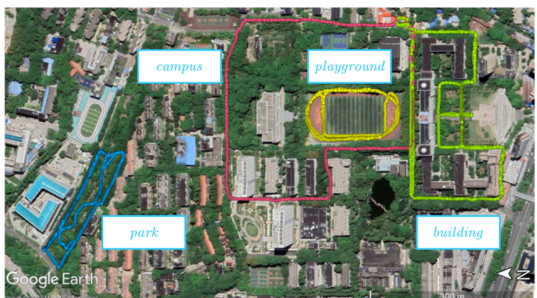

Fig. 6. Testing scenes in the *Robot* dataset. Different colors denote different sequences.

The proposed FF-LINS is implemented using C++ and the robot operating system (ROS). The multi-thread technology is adopted in FF-LINS. The sliding-windows size $n$ is set to 10 to reduce the computational complexity. We assume that the LiDAR-IMU extrinsic and time-delay parameters are all uncalibrated on these datasets. The state-of-the-art (SOTA) tightly-coupled LiDAR-inertial navigation systems LiLi-OM (without loop closure) [18], LIO-SAM (without loop closure) [12], and FAST-LIO2 [19] are employed for comparison. We adopt FF-LINS-WO (without the online calibration) to evaluate the impact of the online calibration of the LiDAR-IMU extrinsic and time-delay parameters. All the systems are run in real-time on a desktop PC (AMD R7-3700X) under the framework of ROS.

### B. Evaluation of the Accuracy

#### 1) Public LiLi-OM Dataset

There is no ground truth in the *LiLi-OM* dataset, and we do not have the end-to-end reference. Hence, the GPS positioning results (with meter-level accuracy) at the starting and ending points are adopted to calculate the starting-ending distance. We also calculate the starting-ending distances of the employed LiDAR-inertial navigation systems. Hence, the distance errors on the *LiLi-OM* dataset are shown in Table I. It should be noted that such an evaluation is not accurate, and thus we can only qualitatively analyze the results.

According to the results in Table I, the proposed FF-LINS yields comparable accuracy to the SOTA methods. FAST-LIO2 exhibits degraded accuracy on *Schloss-2*, as its distance error is far larger than other systems. The distance errors are all meter-level on *Schloss-1* and *Schloss-2*, except for FAST-LIO2, and thus LiLi-OM, LIO-SAM, and FF-LINS achieve the same accuracy. However, FF-LINS yields the best accuracy on *East*, which is the longest sequence. The improvement benefits from the robust INS-centric architecture and the consistent state estimation. Besides, the online calibration of the LiDAR-IMU extrinsic and time-delay parameters may further improve the system consistency and thus improve the accuracy.

#### 2) Public R3LIVE Dataset

In *R3LIVE* datasets, we have the end-to-end reference for quantitative evaluation. We fail to run LiLi-OM on the *R3LIVE* dataset, as LiLi-OM is designed for Livox Horizon rather than Livox AVIA in the *R3LIVE* dataset. We obtain the end-to-end results, as shown in Table II. LIO-SAM fails on *hku_main_building*, mainly because of the few feature points in narrow indoor passages. Nevertheless, direct-based methods FAST-LIO2 and FF-LINS succeed in such environments. FF-LINS achieves superior accuracy than FAS-LIO2 on *hku_main_building* and *hkust_campus_00*. FAST-LIO2 yields the best end-to-end result on *hkust_campus_01* because it can

TABLE I
DISTANCE ERRORS ON THE *LiLi-OM* DATASET

| Error (m) | *Schloss-1* | *Schloss-2* | *East* |
|-----------|-------------|-------------|--------|
| LiLi-OM | 1.36 | 1.27 | 15.43 |
| LIO-SAM | 0.47 | **0.36** | 25.16 |
| FAST_LIO2 | 1.10 | 6.59 | 8.30 |
| FF-LINS | **0.23** | 1.14 | **2.81** |

TABLE II
END-TO-END ERRORS ON THE *R3LIVE* DATASET

| Error (m) | *hku_main_building* | *hkust_campus_00* | *hkust_campus_01* |
|-----------|---------------------|-------------------|-------------------|
| LIO-SAM | Failed | 3.29 | 20.82 |
| FAST-LIO2 | 2.50 | 3.69 | **0.14** |
| FF-LINS-WO | 12.18 | 14.16 | 17.14 |
| FF-LINS | **1.20** | **2.41** | 2.51 |

FF-LINS-WO denotes the configuration without the online calibration of the LiDAR-IMU extrinsic and time-delay parameters in section IV.C.

TABLE III
ARE AND ATE ON THE *ROBOT* DATASET

| ARE / ATE (deg / m) | *campus* | *building* | *playground* | *park* |
|---------------------|----------|------------|--------------|--------|
| FAST-LIO2 | 3.55 / 4.42 | 3.13 / 3.12 | 2.84 / 1.59 | 3.24 / 4.00 |
| FF-LINS-WO | 2.45 / 2.17 | 2.23 / 2.24 | 2.55 / 1.79 | 2.40 / 2.08 |
| FF-LINS | **0.41 / 1.51** | **0.65 / 1.90** | **0.77 / 1.27** | **0.90 / 1.44** |

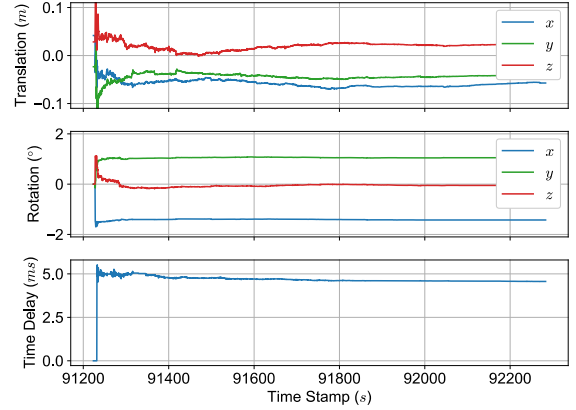

Fig. 7. Estimated LiDAR-IMU extrinsic and time-delay parameters on *R3LIVE-hkust_campus_00*.

match its prebuilt map due to the frame-to-map association. As FAST-LIO2 may also drift without the prebuilt map, such frame-to-map matching will result in a large jump in the trajectory. Besides, *hkust_campus_00* and *hkust_campus_01* are collected in the same testing scenes. FAST-LIO2 exhibits different results on the two sequences, while the proposed FF-LINS yields a similar accuracy. Hence, the result for FF-LINS on *hkust_campus_01* is not so-called bad. The results demonstrate the proposed FF-LINS with the frame-to-frame association is more robust. Besides, FF-LINS is more accurate in terms of consistency in different sequences on *R3LIVE*, *i.e.* *hkust_campus_00* and *hkust_campus_01*.

*3) Private Robot Dataset*

We fail to run LiLi-OM and LIO-SAM on the *Robot* dataset. As Livox Mid-70 only contains one scanning line, few feature points can be extracted, which is terrible for feature-based systems like LiLi-OM and LIO-SAM. The absolute rotation error (ARE) and absolute translation error (ATE) are adopted for quantitative evaluation. Table III indicates that FF-LINS exhibits superior accuracy than FAST-LIO2 on all four sequences. The results may be the sparse single LiDAR frame of Livox Mid-70, resulting in fewer frame-to-map associations than for FAST-LIO2. Hence, FAST-LIO2 degrades accuracy in the *Robot* dataset, especially when large motions occur, which may result in few LiDAR frame-to-map measurements. In contrast, with the INS-centric architecture, the keyframe point-cloud maps are built with several LiDAR frames to construct the frame-to-frame association in FF-LINS. In other words, the proposed frame-to-frame association is more robust. Besides, the INS information is fully utilized in FF-LINS, and the LiDAR-IMU extrinsic and the time-delay parameters are all estimated and calibrated online. Hence, FF-LINS can achieve more consistent state estimation and thus can perform higher navigation accuracy.

*C. The Impact of the Online Calibration*

FF-LINS-WO is adopted to evaluate the impact of the online calibration of the LiDAR-IMU extrinsic and time-delay parameters. Table II shows that FF-LINS-WO indicates significant accuracy degradation without the online calibration on the *R3LIVE* dataset. The reason is that the rotation parts of the LiDAR-IMU extrinsic parameters are relatively large. Specifically, the angles w.r.t the $x$ and $y$ axes (the horizontal attitude angles) are larger than 1.0 degrees, as depicted in Fig. 7. The horizontal attitude angles, *i.e.* the roll and pitch angles, are observable terms due to the gravity [10]. Thus their impacts are much more significant.

Moreover, all LiDAR-IMU extrinsic parameters converge, and even the tiny time-delay parameter converges. The results in Fig. 7 demonstrate that the proposed FF-LINS is consistent in state estimation; thus, these parameters can be effectively estimated. It should be noted the estimated parameters are almost the same on different sequences within a dataset, which proves that FF-LINS is consistent once again.

The results in Table III also demonstrate that the online calibration can notably improve the system accuracy on the *Robot* dataset, especially the rotation accuracy. The reason is that the angle w.r.t the $z$ axis (the yaw angle) of the LiDAR-IMU extrinsic parameters is larger than 2.0 degrees on the *Robot* dataset, according to our analyses. The yaw angle is an unobservable term [10], and thus its impact should be limited, as the translation accuracy only degrades a little.

*D. Running time analysis*

The average running times of FF-LINS on the *Robot* dataset are shown in Table IV. The LiDAR frame preprocessing costs about 0.6 ms per frame. The average interval of the keyframes is determined by the motions of the robot. The FGO times vary on different datasets, as the number of valid frame-to-frame associations may be notably different. According to the results in Table IV, the proposed FF-LINS can perform real-time

| Time (ms) | *campus* | *building* | *playground* | *park* |
|---|---|---|---|---|
| Keyframe interval | 300 | 300 | 310 | 370 |
| Frame-to-frame assoaciation | 2.7 | 2.6 | 2.7 | 2.7 |
| Factor graph optimization | 38.5 | 37.5 | 45.6 | 32.6 |

navigation. According to our statistics, FF-LINS can run at $4\times\sim6\times$ speeds on the desktop PC (AMD R7-3700X), exhibiting superior real-time performance.

## V. CONCLUSIONS

This letter proposes a frame-to-frame solid-state-LiDAR-inertial state estimator, which can achieve robust and consistent navigation in challenging environments. The proposed LiDAR measurement model can provide a relative pose constraint by constructing the direct frame-to-frame data association. Hence, the inconsistency problem in LiDAR-inertial navigation systems due to the frame-to-map association has been solved. The LiDAR-IMU extrinsic and time-delay parameters can be effectively estimated and calibrated online with the consistent state estimator. Besides, we do not need to extract feature points or segment and track plane points, significantly improving the real-time performance.

The proposed LiDAR frame-to-frame measurement model can be seamlessly incorporated into a multi-sensor fusion navigation system with absolute-positioning sensors, such as the GNSS and the high-precision map. Besides, the proposed method provides an effective solution for offline LiDAR-IMU calibrations. In addition, the frame-to-frame association can also be utilized for large-scale and consistent mapping by incorporating loop closure. One of the key points is that the accumulated point clouds are used to build the keyframe point-cloud map for the frame-to-frame associations with the INS-centric architecture. Hence, the proposed FF-LINS is not limited to the Livox LiDARs. It can be applied to other LiDARs, such as MEMS and flash solid-state LiDARs, and even 3D-spinning LiDARs.